\pdfoutput=1

\documentclass[11pt]{article}

\usepackage[svgnames]{xcolor}
\usepackage[final]{acl}

\usepackage{times} 
\usepackage{latexsym}

\usepackage[T1]{fontenc}

\usepackage[utf8]{inputenc}

\usepackage{microtype}

\usepackage{inconsolata}

\usepackage{graphicx}

\usepackage{tabularx} 
\usepackage{longtable}
\usepackage{booktabs} 
\usepackage{multirow}
\usepackage{multicol}

\usepackage{subcaption} 
\usepackage{float} 
\usepackage{stfloats} 

\usepackage{textcomp} 

\definecolor{myviolet}{HTML}{e6b3ff}
\definecolor{myorange}{HTML}{ffd9b3}

\definecolor{mygreen}{HTML}{e4f6e4}
\definecolor{mypurple}{HTML}{ddd1f6}

\title{Lost in Variation? \\ Evaluating NLI Performance in Basque and Spanish Geographical Variants}

\author{Jaione Bengoetxea  \and
  Itziar Gonzalez-Dios  \and Rodrigo Agerri \\ 
  HiTZ Center - Ixa, University of the Basque Country UPV/EHU \\
  \texttt{\{jaione.bengoetxea,itziar.gonzalezd,rodrigo.agerri\}@ehu.eus} \\}

\begin{document}
\maketitle

\begin{abstract}

In this paper, we evaluate the capacity of current language technologies to understand Basque and Spanish language varieties. We use Natural Language Inference (NLI) as a pivot task and introduce a novel, manually-curated parallel dataset in Basque and Spanish, along with their respective variants. Our empirical analysis of crosslingual and in-context learning experiments using encoder-only and decoder-based Large Language Models (LLMs) shows a performance drop when handling linguistic variation, especially in Basque. Error analysis suggests that this decline is not due to lexical overlap, but rather to the linguistic variation itself. Further ablation experiments indicate that encoder-only models particularly struggle with Western Basque, which aligns with linguistic theory that identifies peripheral dialects (e.g., Western) as more distant from the standard. All data and code are publicly available.\footnote{\url{https://huggingface.co/datasets/HiTZ/XNLIvar}}

\end{abstract}

\section{Introduction}\label{sec:intro}

Sociolinguistics examines language variation in relation to various regional, contextual, or social factors. During the 70s and 80s, sociolinguist William Labov highlighted the social aspect of language, and his work on rule-governed language variation has thereby legitimized non-standard language and transformed the study of sociolinguistics. For example, \citet{Labov2006} noted that ``the linguistic behavior of individuals cannot be understood without knowledge of the communities that they belong to''. Thus, \emph{variation} is an intrinsic characteristic of language, influenced by factors such as gender, age, socio-economics, or geographical location. In fact, humans make no distinction when processing their own dialect or the standard variant.

In this regard, \citet{coseriu1956geografia} offered a systematic typology of language variation, based on the following three types: 
(i) diatopic variation, or geographical variation such as dialects, (ii) diastratic variation, or speech of different societal groups, and (iii) diaphasic variation, or speech changes depending on the communicative environment. 

In this paper, we focus on geographical variation in Basque, a low-resource language isolate with around 1 million speakers that is still undergoing a normalization process (started in 1968), and in Spanish, a higher-resourced language whose standardization process started in the 18th century with around 600 million speakers worldwide.


Recent developments in Artificial Intelligence (AI) and Natural Language Processing (NLP) have underscored the significance of social factors in language for NLP systems, as noted by \citet{hovy-yang-2021-importance}. This indicates the importance of developing NLP technology that not only processes standard language but also variations, as this would alleviate any potential language-based discrimination by providing more linguistically-inclusive resources.

\begin{table}[!t]
\begin{small}

    \centering
    \small
    \begin{tabular}{p{0.22\textwidth}|p{0.2\textwidth}} \toprule
    \multicolumn{2}{c}{\textbf{Basque}} \\ \toprule
    \multicolumn{1}{c}{Standard} & \multicolumn{1}{c}{Variation} \\ \midrule
     Zeharo \colorbox{myorange}{hunkituta} gelditu nintzen ezusteko agur \colorbox{myorange}{honekin}  & Asko \colorbox{myviolet}{emoziona} nintzen ezusteko agur \colorbox{myviolet}{horregaz} \\ \bottomrule 
      \multicolumn{2}{c}{\textbf{Spanish}} \\ \toprule
       Me quedé completamente \colorbox{myorange}{conmovido} con \colorbox{myorange}{esta despedida inesperada}  & Me quedé completamente \colorbox{myviolet}{conmovío} con \colorbox{myviolet}{ehta dehpedía inehperá} \\ \bottomrule

       \multicolumn{2}{c}{\textbf{English}} \\ \toprule
       \multicolumn{2}{l}{I was completely surprised by that unexpected goodbye} \\   \bottomrule
    \end{tabular}
    \caption{Example from standard to variation sentences in Basque and Spanish.}
    \label{tab:example}
\end{small}
\end{table}

Although previous work on NLP has primarily focused on standard language, recent research has slightly shifted its attention to the exploration of language variation. For instance, \citet{Zampieri_Nakov_Scherrer_2020} or \citet{joshi2024naturallanguageprocessingdialects} present thorough outlines of variation-inclusive research.  However, due to the lack of data on linguistic variation, most NLP research has focused on a narrow list of languages and their variants, such as Arabic, Indic languages, or German \citep{joshi2024naturallanguageprocessingdialects}. Furthermore, other larger efforts are either based on automatically obtained data or do not provide fine-grained variation distinctions for some widely-spread languages, such as Spanish \cite{FAISAL2024DIALECTBENCHAN,alam-etal-2024-codet}.


Regarding Basque, the few available works have focused on historical dialects \citep{estarrona-etal-2020-dealing} or northern Basque dialects \citep{uria2012hizkeren}. In Spanish, all datasets with linguistic variation have been automatically collected through geolocation techniques \citep{espana-bonet-barron-cedeno-2024-elote, Valentini2024MessIRveAL}. 

In this context, the objective of this paper is to provide the first manually curated variation dataset for Basque and Spanish that captures language variation in real-world usage.  To do so, we introduce XNLIvar, the first variation-inclusive Natural Language Inference (NLI) dataset in Basque and Spanish. An example of an instance from our dataset can be found in Table \ref{tab:example}, which will be used to evaluate current state-of-the-art language models. The main contributions are the following:

\vspace{-0.25cm}
\begin{enumerate}
    \item The first publicly available manually-curated NLI dataset for Basque and Spanish geographic language variations.
\vspace{-0.15cm}
    \item A comprehensive evaluation of encoder-only and decoder-based Large Language Models (LLMs) demonstrates substantially worse performance when processing language variation, particularly in Basque. Detailed error analysis shows that lexical overlap between premise and hypothesis has no impact on the performance drop, which indicates that linguistic variation could be the primary factor for this decrease in accuracy.
\vspace{-0.15cm}
    \item Empirical results suggest that LLM performance with Spanish variants may be attributed to the substantial representation of Spanish-language content in pre-training corpora. Further error analysis suggests that orthographic changes have a substantially negative effect on Spanish language variation processing.
    \vspace{-0.15cm} 
\end{enumerate}

To the best of our knowledge, no work has extensively addressed the automatic processing of language variation of Western and Central Basque dialects in the task of NLI.

\section{Related Work}
\label{sec:related-worlk}

This section presents previous work on language variation in the field of NLP, with a specific focus on Basque language variation. 

\paragraph{Language variation in NLP}

In recent years, there has been an increasing interest in dialects in several fields of NLP, such as dialect identification \citep{ramponi-casula-2023-diatopit}, sentiment analysis \citep{ball-burack-2021}, Machine Translation (MT) \citep{kuparinen-etal-2023-dialect}, and dialogue systems \citep{Alshareef-2020}.

\citet{aepli-sennrich-2022-improving} explored cross-lingual transfer between closely related varieties by adding character-level noise to high-resource data to improve generalization. Moreover,  \citet{ramponi-casula-2023-diatopit} pretrained LLMs for geographic variations of Italian tweets. Finally,  \citet{demszky-etal-2021-learning} showed that BERT models trained on annotated corpora obtained high accuracy for Indian English feature detection.

One of the primary limitations of these studies is the scarcity of available dialectal data. Therefore, research has largely focused on developing resources such as lexicons and dialectal datasets on a small subset of languages: \citet{artemova-plank-2023-low} propose a bilingual lexicon induction method for German dialects using LLMs, while \citet{hassan-etal-2017-synthetic} introduce a synthetic data creation method through embeddings by transforming input data into its dialectic variant.  With respect to language coverage, the Arabic family, due to its relative data availability, has received the most attention, followed by Indic languages, Chinese, and German \cite{joshi2024naturallanguageprocessingdialects}.

\paragraph{Basque language variation}

In dialectology, \citet{zuazu2008} established an extensive and comprehensive descriptive representation of features of modern Basque dialects. In NLP, \citet{estarrona-etal-2020-dealing} worked on a morpho-syntactically annotated corpus of Basque historical texts as an aid in the normalization process. Moreover, \citet{uria2012hizkeren} introduced a corpus of syntactic variation in northern Basque dialects.

Additionally, some dialectal benchmark works have included Basque in their experimentation: both  \citet{alam-etal-2024-codet} and  \citet{FAISAL2024DIALECTBENCHAN} presented benchmarks for MT with northern Basque dialects. 


\paragraph{Spanish language variation}

Several works have dealt with Spanish varieties. For instance, \citet{espana-bonet-barron-cedeno-2024-elote} automatically filtered Open Super-large Crawled Aggregated coRpus (OSCAR) by geolocation into different Spanish variants and performed a stylistic analysis. \citet{Valentini2024MessIRveAL} automatically collected Google queries from several Spanish-speaking countries and provided an Information Retrieval baseline for Spanish varieties. 

Additionally, \citet{lopetegui-etal-2025-common} introduced a Cuban Spanish dataset by collecting geolocated tweets from Twitter. They focused their study on common examples, i.e., instances that can be valid across several dialects. They performed a manual annotation of tweets into \emph{Cuban dialect}, \emph{other dialect}, or \emph{common example}. Similarly, \citet{castillo-lopez-etal-2023-analyzing} collected tweets from European and Latin American geolocations and annotated them for hate speech.

\section{Data}\label{sec:data}


In this work, we introduce a novel dataset, \textbf{XNLIvar}, that expands the XNLI framework by generating various dialectal variations for both Basque and Spanish languages. We choose NLI as an evaluation framework because it is considered to be a general benchmark for evaluating language understanding, which requires dealing with semantic relationships, logical implications, world knowledge, and contextual nuances \cite{williams-etal-2018-broad,conneau-etal-2018-xnli,artetxe-etal-2020-translation}, including figurative language \cite{naik-etal-2018-stress,stowe-etal-2022-impli,liu-etal-2022-testing,sanchezbayona2024meta4xnli}. NLI is a fundamental NLP task that involves classifying the logical relationship between two segments (a premise and a hypothesis) as one of three categories: entailment (the hypothesis logically follows from the premise), contradiction (the hypothesis contradicts the premise), or neutral (the hypothesis neither follows from nor contradicts the premise). The most popular dataset is the English \textbf{MultiNLI} \cite{williams-etal-2018-broad}.



\textbf{XNLI} \citep{conneau-etal-2018-xnli} provides an extension of MultiNLI in 15 languages, among them Spanish (\textbf{XNLIes}). The training set for each language was created by translating the original MNLI data. However, as the test partition of MNLI is not public, \citet{conneau-etal-2018-xnli} collected 7,500 English examples via crowdsourcing, which were then professionally translated to create the development (2500 instances) and the test (5K instances) splits of XNLI. This parallel multilingual corpus has facilitated crosslingual NLI research beyond English-centric approaches by exploring model-transfer, translate-train, and translate-test techniques to alleviate the lack of annotated training data in a given target language \cite{artetxe-etal-2020-translation, artetxe-etal-2022-corpus}.


\textbf{XNLIeu} is a professionally translated version of the English XNLI set into Basque \cite{heredia-etal-2024-xnlieu}, a language not included in the original XNLI dataset. Additionally, we also use \textbf{XNLIeu\textsubscript{native}}, an NLI dataset generated by collecting native Basque premises and hiring Basque annotators to create three hypotheses per premise \cite{heredia-etal-2024-xnlieu}. The experimental results from XNLIeu demonstrate that NLI systems exhibit significant performance sensitivity to disparities between training and testing data distributions, highlighting the critical role of data provenance \citep{artetxe-etal-2020-translation, voansky-2013}.


\subsection{XNLI with Geographic Variants}\label{sec:rewrite}


To investigate the impact of language variation via evaluation in NLI, we developed two novel \textbf{XNLI variants} datasets encompassing Basque and Spanish geographic-based linguistic variations, namely, \textbf{XNLIeu\textsubscript{var}} and \textbf{XNLIes\textsubscript{var}}. The methodology involved a language adaptation phase to ensure the incorporation of variant diversity within the data. These two variant datasets were developed taking \textbf{XNLIeu\textsubscript{native}} as a starting point for dialectal augmentation due to its authentic representation of Basque language patterns and its suitable scale for manual paraphrasing.



The adaptation process was the same for Basque and Spanish, including native speakers as linguistic informants for variant transformation.  We wanted to analyze the variation that naturally occurs among native speakers, employing minimally restrictive parameters to capture authentic dialectal features. Thus, informants were instructed to perform dialectal adaptations of source sentences, with allowance for modifications across multiple linguistic dimensions, including lexical, grammatical, phonetic, and orthographic alterations. The full adaptation guidelines are detailed in Appendix  \ref{sec:appendix1}.


\paragraph{XNLIeu\textsubscript{var}}


Twelve native Basque speakers were recruited from diverse geographical regions. All participants possessed expertise in NLP and held university degrees in either Linguistics, Computer Science, or Engineering. Each participant was tasked with reformulating approximately 20 brief sentences, with the resulting adaptations categorized according to three major dialectal variants: Western, Central, and Navarrese. To facilitate cross-dialectal comparison, a subset of 10 identical sentences was assigned to more than one annotator, enabling parallel dialectal representations. The demographic and professional characteristics of the annotators, including age, gender, and educational background, are detailed in Appendix \ref{sec:appendix2}.



It should be noted that during data collection, a single annotator generated two types of variants for each sentence, including both dialectal variations and allocutive agreement forms in Basque. The allocutive system in Basque requires morphological marking of the addressee's gender (masculine/feminine) within the verbal form. Consequently, \textbf{XNLIeu\textsubscript{var}} exhibits a higher instance count (894) compared to the original \textbf{XNLIeu\textsubscript{native}} dataset (621), as shown in Table \ref{tab:train-test-data}.

In terms of dialect distribution, 592 instances correspond to the Central dialect, usually associated with the province of Gipuzkoa, 240 instances to the Western dialect (West Gipuzkoa and Biscay), and just 63 instances to the Navarrese dialect, comprising 7\% of the data. Thus, the Navarrese dialect is clearly under-represented in our data. 


\paragraph{XNLIes\textsubscript{var}}

XNLIeu\textsubscript{native} was automatically translated into Spanish using Claude 3.5 Sonnet\footnote{\url{https://www.anthropic.com/news/claude-3-5-sonnet}}, generating the \textbf{XNLIeu2es\textsubscript{native}} dataset and facilitating the creation of a parallel corpus for Basque and Spanish texts with their respective variants. Quality verification was conducted through manual review of the machine-generated translations, making sure that they constituted an authentic representation of Spanish language patterns.  Finally, the translated corpus was provided to Spanish-language annotators for variant-specific adaptation.



The adaptation task involved six independent annotators, each assigned a set of 50 sentences for dialectal adaptation into their respective Spanish variants. They represented four distinct geographical locations: Cuba, Ecuador, Spain, and Uruguay. Two annotators from Spain performed adaptations into separate dialectal variants (Andalusian and Tenerife), resulting in a total of five Spanish dialectal variations in the final dataset. The demographic and professional characteristics of the annotators, including age, gender, and educational background, are documented in Appendix \ref{sec:appendix2}.

It is worth noting that some annotators found it difficult to add dialectal features to the standard sentences. This could be due to the high number of common examples in Spanish varieties \cite{lopetegui-etal-2025-common, zampieri-etal-2024-language}. In other words, the distinctions between Spanish varieties tend to be more homogeneous and thus contain less variation compared to Basque (Section \ref{sec:error_anal}).


Similar to the Basque adaptation, multiple dialectal variants were documented by some annotators. These variants exhibited phonological phenomena such as word-final /s/ deletion (e.g., {\it digamos} $\to$ {\it digamo}) and /s/ to /j/ substitution in word-final position (resulting in {\it digamoj}). Thus,  XNLIes\textsubscript{var} contains 666 examples, representing a marginally higher count than the base dataset.

Table \ref{tab:train-test-data} provides an overview of the datasets used for experimentation, including our newly generated \textbf{XNLIvar}, consisting of XNLIeu\textsubscript{var} and XNLIes\textsubscript{var}.

\section{Experimental settings}\label{sec:setting}

\begin{table}[!t]
\begin{small}
    
\centering
\begin{tabular}{lr} \\  \toprule
\multicolumn{2}{c}{\textbf{Train}} \\ \toprule
Dataset       & Instances       \\ \toprule
MNLI          &        392k    \\
MNLIeu        &     392k       \\
MNLIes        &       392k     \\ \toprule
\multicolumn{2}{c}{\textbf{Test}}  \\ \toprule
    XNLIeu &  5010 \\
    XNLIes & 5010 \\ \midrule
    XNLIeu\textsubscript{native}  & 621 \\ 
    XNLIeu2es\textsubscript{native}  & 621\\ \midrule
    XNLIeu\textsubscript{var}  & 894 \\
    XNLIes\textsubscript{var}  & 666\\ 
    \bottomrule 
\end{tabular}
\caption{Datasets used for training and testing.}
\label{tab:train-test-data}

\end{small}
\end{table}

Empirical research was based on the aforementioned datasets to evaluate the impact of dialectal variation on NLI performance. 

\paragraph{Discriminative experiments} Table \ref{tab:config-encoders} illustrates the experiments performed using encoder-only Transformer models and the datasets specified in Table \ref{tab:train-test-data}. 

\vspace{-0.20cm}
\begin{itemize}
    \item \textbf{Model transfer:} The train split of the original MNLI (English) is used to fine-tune multilingual encoder models. Evaluation is performed on the test sets for Basque and Spanish specified in Table \ref{tab:train-test-data}.
    \vspace{-0.25cm}
    \item \textbf{Translate-train:} The MNLI training is automatically translated into Basque and Spanish (MNLI\textsubscript{eu} and MNLI\textsubscript{es}); multilingual and monolingual encoders are then fine-tuned using the translated training data and evaluated in each of the target languages. 
\vspace{-0.25cm}
    \item \textbf{Translate-test:} Tests in the target languages are translated into English and evaluated using the MNLI fine-tuned encoders (in English). 
\end{itemize}
\vspace{-0.20cm}

\begin{table}[!htbp]
\begin{small}

\centering
\begin{tabular}{ccc}
\toprule
\textbf{Configuration} & \textbf{Train} & \textbf{Test}  \\ \toprule

Model transfer & English  & Target language
   \\ \midrule

Translate-train  & Target language  & Target language
  \\ \midrule
  
Translate-test  & English & Target → English  \\ \bottomrule
\end{tabular}
\caption{Discriminative model configurations and data. \textbf{→:} Translated to. }
\label{tab:config-encoders}

\end{small}
\end{table}

\begin{table*}
\begin{small}

\centering
    \resizebox{\textwidth}{!}{
    \begin{tabular}{lrrr|rrr|rrr} \toprule 
    & \multicolumn{9}{c}{\textbf{Basque}} \\ \midrule 
     & \multicolumn{3}{c|}{Model transfer} & \multicolumn{3}{c|}{Translate-train} & \multicolumn{3}{c}{Translate-test} \\ \midrule
    
      &  XNLIeu &  XNLIeu\textsubscript{native} &  XNLIeu\textsubscript{var}  &  XNLIeu &  XNLIeu\textsubscript{native} &  XNLIeu\textsubscript{var} &  XNLIeu &  XNLIeu\textsubscript{native} &  XNLIeu\textsubscript{var} \\ \midrule
      XLM-RoBERTa large &  80.00  &  72.09  &  68.24 &  \textbf{83.42} & 75.63  &  \textbf{73.21}   &  - & \textbf{75.85}  &  71.63  \\ 
      mDeBERTa&  78.95  &  70.21  &  67.26 &  81.42 & 72.14  &  69.77  &  - & 72.68  &  70.28     \\ 
      
    \midrule 

      & \multicolumn{9}{c}{\textbf{Spanish}} \\ \midrule 

      &  XNLIes  &  XNLIeu2es\textsubscript{native}  &  XNLIes\textsubscript{var}  &  XNLIes  &  XNLIeu2es\textsubscript{native}  &  XNLIes\textsubscript{var} &  XNLIes  &  XNLIeu2es\textsubscript{native}  &  XNLIes\textsubscript{var} \\ \midrule
       XLM-RoBERTa large &  83.05  &  74.02  &  73.07 &  \textbf{84.69} & \textbf{74.61}  &  \textbf{73.72}  &  - & 73.86  &  71.77  \\
      mDeBERTa &  82.02  &  74.13  &  71.57 &  83.27 & 72.25  &  70.77  &  - & 72.30  &  69.89  \\ 
    \bottomrule
    
    \end{tabular}
    }
        \caption{Accuracy results for Basque and Spanish discriminative experiments.}
    \label{tab:discriminative-all}
\end{small}
\end{table*}
\begin{table*}
\begin{small}
\centering
    \resizebox{\textwidth}{!}{
    \begin{tabular}{lrrrrr|rrrrr} \toprule 
   
     & \multicolumn{10}{c}{\textbf{Basque}} \\ \midrule 
     & \multicolumn{5}{c|}{\textbf{Llama-3.1-Instruct-70B}} & \multicolumn{5}{c}{\textbf{Gemma-2-it-27B}}  \\ \midrule
    
       &  nli-zero  &  nli-few  &  qa-zero  &  qa-few  &  chain &  nli-zero  &  nli-few  &  qa-zero  &  qa-few  &  chain  \\ \midrule
       XNLIeu &  33.65 & 53.17 & 33.31 & 54.89 & \textbf{55.25} & 61.10 & 62.81 & 61.84 & \textbf{65.27} & 58.28 \\
       
      XNLIeu\textsubscript{native} &  38.81 & 56.68 & 39.61 & 58.61 &	\textbf{60.71} & 64.90 & 66.67 & 65.70 & \textbf{68.28}	& 66.99 \\ 
      
      XNLIeu\textsubscript{var} & 33.78 & 48.66 & 31.54 & \textbf{50.11} & 49.22 & 57.61	& 60.96 & 57.49 & \textbf{61.52} & 58.05 \\ 
      \toprule
 
       & \multicolumn{10}{c}{\textbf{Spanish}} \\ \midrule 

     & \multicolumn{5}{c|}{\textbf{Llama-3.1-Instruct-70B}} & \multicolumn{5}{c}{\textbf{Gemma-2-it-27B}}  \\ \midrule
    
       &  nli-zero  &  nli-few  &  qa-zero  &  qa-few  &  chain &  nli-zero  &  nli-few  &  qa-zero  &  qa-few  &  chain  \\ \midrule
       XNLIes & 54.65 & 62.18	& 51.54 & 65.69 & \textbf{73.97}  & 66.75 & 71.28 & 70.52 & \textbf{73.05} & 68.88 \\
       
      XNLIeu2es\textsubscript{native} & 62.96 & 62.48 & 62.16 & 70.69 &	\textbf{77.29} & 71.50 & 72.62 & 73.91 & 73.43 & \textbf{76.97}  \\ 
      XNLIes\textsubscript{var} & 59.42 & 62.32 & 54.27 & 69.24 & \textbf{75.52}& 70.37 & 72.30 & 72.79 & 72.14 & \textbf{74.56} \\
      \toprule
   \end{tabular}
   }

\caption{Results with generative LLMs.}
\label{tab:generative-all}
\end{small}
\end{table*}

Summarizing, training is always done with MNLI, either in its original English form or using the automatically translated versions to Basque and Spanish. Moreover, there are three different test data types: (i) XNLI test data professionally translated into the target languages (XNLIeu, XNLIes) (ii) the manually created native Basque data and its translation to Spanish (XNLIeu\textsubscript{native}, XNLIeu2es\textsubscript{native}) and, (iii) the native datasets adapted to different variations for each of the target languages (XNLIeu\textsubscript{var}, XNLIes\textsubscript{var}).



We employed two multilingual encoder-only language models for our target languages: XLM-RoBERTa large \cite{conneau-etal-2020-unsupervised} and mDeBERTa \citep{debertav3}. The hyperparameter configuration followed \citet{heredia-etal-2024-xnlieu}, implementing differential learning rates of 5e-5 and 10e-6 for BERT and RoBERTa architectures, respectively. All other parameters were maintained at their default values. The training process consisted of 10 epochs across all model configurations.

\paragraph{Generative experiments} We experimented with generative LLMs to evaluate the decoders' ability to perform NLI when language variation is present. We started with a zero-shot setting, where we prompt LLMs to identify the NLI relation.



We also evaluated alternative prompting methodologies, specifically, few-shot and Chain of Thought (CoT) approaches. The few-shot prompt implemented a single example for each classification category. The CoT methodology incorporated detailed task-specific contextual information alongside a single example for each label.


To further evaluate the linguistic comprehension capabilities of LLMs with respect to Basque and Spanish variants, we implemented an alternative methodological approach by transforming the NLI task into a Question-Answering (QA) setting. In this experimental configuration, the input prompt was restructured as a question to be answered by the LLM, with the three possible answers based on the NLI inference labels. Zero-shot and few-shot prompting strategies kept the same. The complete set of prompt templates used across all task formulations is available in Appendix \ref{sec:app_promp}.


We selected Llama-3.1-Instruct (8B and 70B versions) \cite{Dubey2024TheL3} and Gemma 2 instruct (9B and 27B versions) \cite{Mesnard2024GemmaOM} due to their strong performance in both Basque and Spanish languages\footnote{\url{https://hf.co/spaces/la-leaderboard/la-leaderboard}} \cite{etxaniz-etal-2024-latxa,figueras2025truthknows}. In the next section we focus on the results obtained by the larger LLMs (performances with smaller LLMs in Appendix \ref{sec:addicional-generative-results}).

\section{Results}
\label{sec:results}

We first report the results obtained in the discriminative settings, while in Section \ref{sec:gen-exp}, we discuss the results of in-context learning with LLMs.

\subsection{Discriminative Experiments} 

By looking at the results reported in Table \ref{tab:discriminative-all}, the empirical results demonstrate a significant performance degradation when comparing XNLIeu and XNLIes against the native and variation datasets. This observation aligns with existing literature documenting the adverse effects of train-test distribution shifts in cross-lingual settings \cite{artetxe-etal-2020-translation, voansky-2013}. When comparing native and variation data results, where the only difference is the presence of dialectal data, we see a decrease in results. Therefore, results show that language models perform worse when variants are included in the NLI task.

By doing a cross-configuration analysis, we see that for Basque, the best results are obtained with XLM-RoBERTa in the translate-train for XNLIeu (83.42) and XNLIeu\textsubscript{var} (73.21), while for XNLIeu\textsubscript{native} (75.85), the train-test is superior. Overall, the empirical results demonstrate that the translate-train approach with XLM-RoBERTa yielded the best overall performance for Spanish and Basque. This suggests that training and evaluating in the target language constitutes the optimal method, irrespective of whether the data includes standard or variation-inclusive linguistic content.

Regarding \emph{native} and \emph{variant} results, analysis reveals that Spanish consistently outperforms Basque across all settings and evaluation datasets, demonstrating greater resilience to linguistic variation. In fact, while Spanish accuracy drops minimally (less than 1 percentage point in most cases), Basque performance suffers a higher decrease, with model-transfer and translate-test approaches showing an approximately 4-point drop and translate-train a 2.5-point drop. This highlights a sharper impact of variation on Basque performance.

These results show that when English is the source training data, model-transfer provides competitive results for a high-resource, structurally similar language such as Spanish, while for a low-resource and morphologically different language such as Basque, the data-transfer (translate-train) strategy remains preferable \cite{agerri-etal-2020-give,artetxe-etal-2020-translation,garcia2022model}. 

Overall, the consistently lower performance observed in Basque relative to Spanish across all evaluation conditions can be attributed to three key factors: (i) Basque's agglutinative morphological structure, (ii) its classification as a language isolate, and (ii) reduced Basque language representation in the models' pre-training data \cite{agerri-etal-2020-give,etxaniz-etal-2024-latxa}.






Finally, we also experimented with two Basque monolingual models, RoBERTa-Euscrawl \cite{artetxe-etal-2022-corpus} and BERTeus \cite{agerri-etal-2020-give}, in the translate-train setting. However, while competitive, their results did not outperform those obtained by XLM-RoBERTa large. Further details can be found in Appendix \ref{sec:monolingual-discriminative}.

\begin{figure*}[ht]
    \centering
    \begin{subfigure}[t]{0.48\textwidth}
        \centering
        \includegraphics[width=\linewidth]{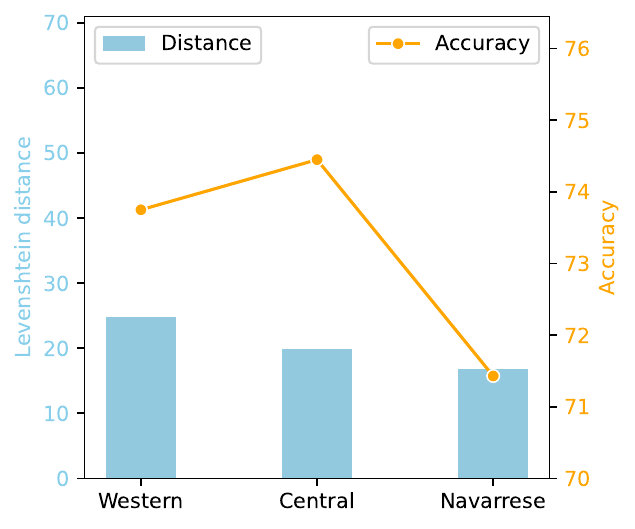}
        \caption{Basque results}
        \label{fig:dialect-distance-Basque}
    \end{subfigure}
    \hfill
    \begin{subfigure}[t]{0.48\textwidth}
        \centering
        \includegraphics[width=\linewidth]{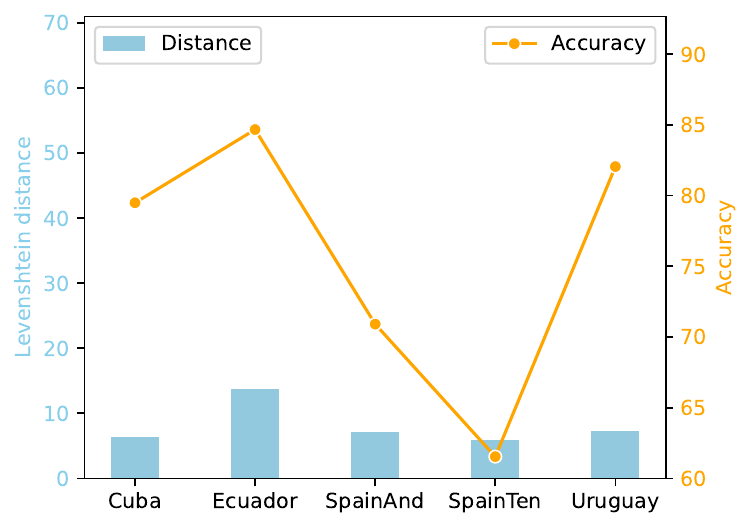}
        \caption{Spanish results}
        \label{fig:es-dialect-distance-Spanish}
    \end{subfigure}
    \caption{Standard to dialectal Levenshtein distance vs accuracy of best discriminative models.}
    \label{fig:accuracy_vs_distance}
\end{figure*}

\subsection{Generative Experiments}
\label{sec:gen-exp}

Table \ref{tab:generative-all} presents the evaluation results for LLMs in the task on variation-inclusive NLI with the largest LLMs tested, namely, Llama-3.1-Instruct-70B and Gemma-2-it-27B.


A first observation reveals a significant performance degradation across all evaluated LLMs when transitioning from standard datasets (XNLIeu\textsubscript{native} and XNLIeu2es\textsubscript{native}) to their variant counterparts (XNLIeu\textsubscript{var} and XNLIes\textsubscript{var}). This suggests a substantial limitation in the capacity of LLMs to process and comprehend linguistic variations within the task. The results also indicate that including examples in the prompt engineering process yields positive effects (in the qa-few and CoT methodologies). Notably, for Spanish, the CoT approach demonstrates superior performance compared to XLM-RoBERTa large on XNLIeu2es\textsubscript{var} and XNLIes\textsubscript{var} datasets.


Concerning Basque, the experimental results demonstrate that Gemma-2 exhibits better performance compared to Llama-3.1. Moreover, for XNLIeu\textsubscript{var} Gemma's optimal performance (61.52) experiences a reduction of 6.5 percentage points relative to the standard XNLIeu\textsubscript{native} (68.28). In contrast, Llama-3.1 exhibits a more substantial decline of 10 percentage points in XNLIeu\textsubscript{var} performance. These findings indicate that Gemma maintains greater robustness against linguistic variation compared to Llama-3.1. 

For Spanish, CoT prompting generally yields the highest accuracy. The variation-inclusive evaluation dataset (XNLIes\textsubscript{var}) produces results very close to those of XNLIes\textsubscript{native}, with Llama 3.1 achieving 75.77 and 77.29, and Gemma 2 reaching 74.56 and 76.56, respectively. Despite this closeness, linguistic variation still causes a drop in accuracy. Overall, Llama 3.1 performs slightly better than Gemma 2, though the difference is minimal.

The empirical evidence obtained from these analyses of Basque and Spanish language understanding indicates that LLMs exhibit significant limitations in their capacity to comprehend linguistic content when confronted with dialectal and geographical variations.

\section{Error analysis}\label{sec:error_anal}

This section presents a quantitative error analysis to evaluate XLM-RoBERTa large's performance with respect to variation-inclusive evaluation data. 

\paragraph{Dialect to standard distance}


The Levenshtein distance metric, which quantifies the minimum number of single-character operations (insertions, deletions, or substitutions) necessary for string transformation, was computed between dialectal and standard sentences. The analysis of distance results demonstrates that Basque dialectal variants (Figure \ref{fig:dialect-distance-Basque}) exhibit significantly greater divergence from the standard form compared to Spanish variants (Figure \ref{fig:es-dialect-distance-Spanish}), which display higher proximity to their standardized counterpart. The observed inter-dialectal variation patterns suggest a more pronounced linguistic differentiation within Basque dialectal systems relative to Spanish dialectal varieties. This emphasizes the difference in variation between languages and highlights the importance of language-specific analysis in the field of language variation processing in NLP.

\paragraph{Accuracy per dialect} We analyzed the accuracy results for each individual dialect class, in order to see if some dialects are more difficult to process than others. The relation between the accuracy for each dialect and the distance from standard to dialect is illustrated in Figure \ref{fig:accuracy_vs_distance}.


In the case of Basque (Figure \ref{fig:dialect-distance-Basque}), we see that, in terms of string distance, the Western dialect is the one that is the most different from the standard, followed by the Central and Navarrese dialects. However, the lowest accuracy is accounted for in the Navarrese dialect, which is the dialect label that seems to be closest to the standard form of language. This could be because of its under-representation in our dataset, as Navarrese examples comprise only 7\% of our data (Appendix \ref{sec:appendix2}). When focusing on Western and Central dialects, it can be observed that, as the distance from standard to dialectal gets higher, accuracy gets lower, suggesting that dialects further from the standard (in our case, Western) are harder to process. The Central dialect being closer to the standard is expected, as it served as the main foundation for the current standard form of Basque.

In fact, according to research in Basque dialectology, peripheral dialects have been found to be more distant from the rest \citep{michelena1981lengua}. This fact has also been corroborated by NLP studies analyzing Basque historical dialects, where Biscayan (Western) and Souletin display the greatest difference \citep{estarrona2023measuring}. Additionally, research has documented the Bizkaian dialect's historical tendency toward linguistic divergence, both from other Basque dialects and from its own earlier forms \cite{zuazuWeb}.


In Spanish variants, Figure \ref{fig:es-dialect-distance-Spanish} shows Ecuador and Uruguay displaying the highest distance values and accuracy scores. Further analysis has shown that adaptations into these two variants mostly include replacing lexical words with alternatives that are more commonly used in those varieties (e.g., \textit{construccion futura > nuevos edificios}), as well as grammatical structures typical of those dialects (e.g., \textit{he podido > pude}). However, standard orthography has been preserved throughout. 

In turn, adaptations into Cuban, SpainAnd and SpainTen variants mostly include phonological or orthography changes (e.g. \textit{misma > mihma, fuerza > fuersa}), which have resulted in lower distance to the standard form of Spanish, but a decrease in accuracy compared to the variants written in standard orthography (Ecuador and Uruguay). This reveals a correlation between standard orthography and high accuracy, and highlights the difficulties of discriminative models to deal with data which includes non-standard orthography. This analysis is illustrated in Appendix \ref{sec:spanish-analysis}. These results match those observed in earlier studies, where orthography variations have also been found to be problematic \citep{delarosa2024modernifa}. Additional results of per-dialect accuracy results are presented in Appendix \ref{sec:accuracy-per-dialect}.

\paragraph{Ablation Tests}

\begin{table}[!t]
\begin{small}
    
\centering
\begin{tabular}{lrrr}
\toprule

\multicolumn{4}{c}{\textbf{Basque}} \\ \midrule

\textbf{Dataset} & \textbf{Instances} & \textbf{Discrimin.} & \textbf{Genera.}\\ \midrule

XNLIeu\textsubscript{native}  & 621  & 75.63 & 68.28 
   \\ \midrule

XNLIeu\textsubscript{var}  & 894  & 73.21 & 61.52
   \\ \midrule


Less-western  & 834  & 73.14 & 60.79
    \\ \midrule

Less-central  & 834  & 72.70 & 61.03
    \\ \midrule

  
No repetitions  & 621 & 71.77 & 60.39
    \\ \toprule

\multicolumn{4}{c}{\textbf{Spanish}} \\ \midrule

XNLIeu2es\textsubscript{native}  & 621  & 74.61 & 77.29
   \\ \midrule

XNLIes\textsubscript{var}  & 666  & 73.72 & 75.52 
   \\ \midrule

No repetitions  & 621  & 73.00 & 77.13
    \\ \bottomrule

\end{tabular}
\caption{Ablation experiments on Basque and Spanish variation data (XNLIeu\textsubscript{var} and XNLIes\textsubscript{var}, respectively). Results obtained using the best discriminative setting (Translate-train XLM-RoBERTa large in Table \ref{tab:discriminative-all}) as well as best generative results for Basque (Gemma-2 qa-few) and Spanish (Llama-3.1 chain) in Table \ref{tab:generative-all}. }
\label{tab:ablation}

\end{small}
\end{table}


As explained in Section \ref{sec:rewrite} and illustrated in Table \ref{tab:train-test-data}, test data in XNLIeu\textsubscript{var} and XNLIes\textsubscript{var} contains duplicated instances in different dialects. In order to see the effect that different types of variation have on accuracy, we have performed some ablation experiments. 


Four different Basque speakers (two Western and two Central) adapted the same 10 sentences, providing us with four distinct versions of those 10 sentences. We used these instances to create two new versions of the dataset, one by removing the repeated sentences from Western variants (\textit{Less-western}), and another one without the repeated instances from the Central one (\textit{Less-central}). Table \ref{tab:ablation} presents accuracy results with these datasets. 

The results show that accuracy is higher when Western-dialect instances are removed (73.14) than when Central instances are excluded (72.70). 


Additionally, we removed all duplicated variant instances from XNLIvar\textsubscript{eu}, resulting in a completely parallel variation dataset to XNLIeu\textsubscript{native} (\textit{No repetitions}), which allows us to calculate whether the results between the standard and the variant versions are statistically significant. As reported in Table \ref{tab:ablation}, accuracy between \emph{No repetitions} and the \emph{standard} substantially decreases (71.77 vs 75.63) for the Basque discriminative experiments. According to a chi-square test of independence, this difference is highly statistically significant ($p$ < .001, df=1). Similar to Basque, all the repeated variant instances from the Spanish variation dataset were removed, obtaining a parallel dataset to XNLIes\textsubscript{native}. Using the \emph{No repetitions} split, a chi-square test of independence establishes that differences with results on XNLIeu2es\textsubscript{native} are highly statistically significant ($p$ < .001,  df=1).


Generative LLMs follow the same trend, with differences in performance being highly statistically significant in both languages  ($p$ < .001,  df=1).




\paragraph{Premise and hypothesis lexical overlap}


\begin{figure*}
\begin{small}
    \centering
    \begin{subfigure}{0.40\textwidth}
        \centering
        \includegraphics[width=\textwidth]{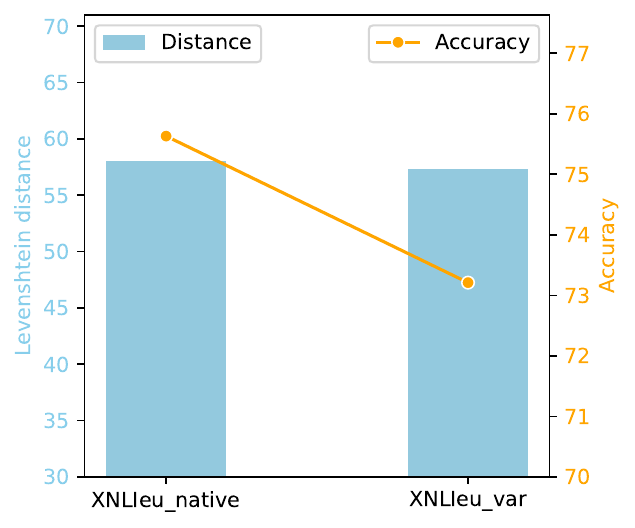}
        \caption{Premise-Hypothesis distance and accuracy for Basque}
        \label{fig:lexical-overlap-eu}
    \end{subfigure}
    \hspace{1cm}
    \begin{subfigure}{0.40\textwidth}
        \centering
        \includegraphics[width=\textwidth]{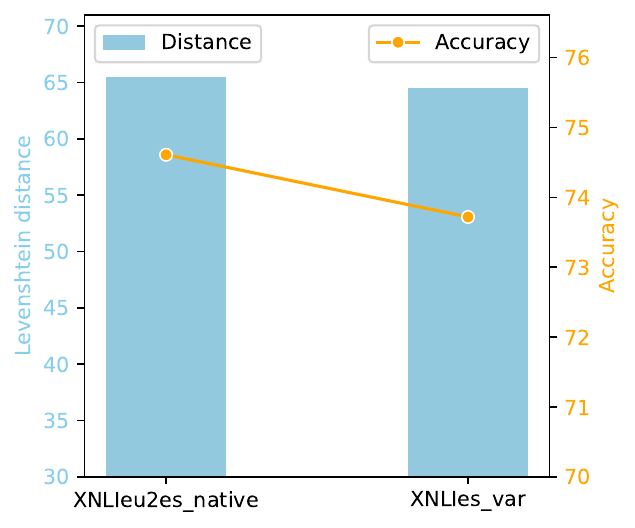}
        \caption{Premise-Hypothesis distance and accuracy for Spanish}
        \label{fig:lexical-overlap-es}
    \end{subfigure}
\caption{Levenshtein distance from premise to hypothesis and accuracy of discriminative models.}
\label{fig:lexical-overlap}
\end{small}
\end{figure*}




To investigate the potential correlation between lexical overlap and accuracy, we measured the Levenshtein distance between premises and hypotheses. The analysis of the data in Figure \ref{fig:lexical-overlap} indicates that lexical overlap remains consistent across standard and dialectal varieties, while a substantial decrease in accuracy was observed in both Basque and Spanish datasets. These findings suggest that while lexical overlap appears to have minimal impact on accuracy metrics, linguistic variation emerges as the significant factor affecting performance. Therefore, the observed pattern implies that dialectal variations, rather than lexical similarities, may be the primary factor of accuracy degradation in this context.




In fact, Figure \ref{fig:lexical-overlap-eu} demonstrates a more pronounced decrease in accuracy for Basque compared to Spanish, underscoring both the critical need to improve Basque representation in multilingual discriminative models and the necessity for additional investigation into language variation processing.

\section{Concluding Remarks}


This paper presents a novel dataset that includes geographical variants of Basque and Spanish. The dataset represents the first documented instance of a manually-curated, variation-inclusive corpus for these languages, facilitating research and evaluation on linguistic variants via NLI. Additional speaker metadata expands its value as a resource for sociolinguistic research on generational and geographical differences in Basque and Spanish. Our investigation involved the empirical evaluation of both discriminative and generative language models across various NLI task configurations.



Results indicate that language models' performance drops when linguistic variation is present. This performance degradation is particularly pronounced in Basque variants, where linguistic variation is higher compared to Spanish variants.  Furthermore, the performance drop intensifies proportionally with the linguistic distance between dialectal variants and their respective standardized forms for Basque, with a higher impact in the Western dialect. This coincides with previously established linguistic theory, which states that some Basque dialects (such as Western) have a historical tendency to distance themselves from the standard. In the case of Spanish, variants with non-standard orthography have shown a significant accuracy drop. Finally, the lexical overlap between premises and hypotheses appears to have minimal impact, suggesting that lower performance is due to linguistic variation.



Future work will involve expanding the dataset to include additional geographical variants of both Basque and Spanish, as well as incorporating other languages. Investigation of variation-inclusive monolingual models represents a promising avenue for future research.










\section*{Limitations}


In this paper, we have focused on geographic variants of language due to their low representation in NLP. We conducted our experiments for a lesser-resourced language, Basque, and a higher-resourced language, Spanish. However, we have only represented some of the variations of these languages, and our variation datasets have been created by 12 speakers for Basque and 6 speakers for Spanish. We tried to include the most representative dialects with different kinds of speakers, but we are aware that all the speakers have linguistic and NLP backgrounds, and laypeople could contribute differently. 

Our empirical findings demonstrate decreased accuracy in natural language inference tasks within our variants dataset. However, generalization of these results requires expansion to include additional linguistic variants and evaluation across a broader range of NLP tasks.


To augment the dataset, we are recruiting speakers from diverse linguistic backgrounds to contribute additional variation data. We further intend to evaluate the performance of NLP tools and LLMs on tasks incorporating dialectal and register variation. 

\section*{Acknowledgments}
This work has been supported by the HiTZ center and the Basque Government (Research group funding IT-1805-22).
Jaione Bengoetxea is funded by the Basque Government pre-doctoral grant (PRE\_2024\_1\_0028).
We also acknowledge the following MCIN/AEI/10.13039/501100011033 projects: (i) DeepKnowledge (PID2021-127777OB-C21) and by FEDER, EU; (ii) DeepMinor (CNS2023-144375) and European Union NextGenerationEU/PRTR.

We acknowledge the contributions of the following people to generate the Basque and Spanish linguistic variants: Ainara Estarrona, Alex Muñoz Alvarado, Ekhi Azurmendi, Irune Zubiaga, Itziar Aldabe, Izaskun Aldezabal, Jon Alkorta, Jose Maria Arriola, Leonel Cosme, María Teresa Martín-Valdibia, Marta Torres Martínez, Mikel Zubillaga, Naiara Perez, Nancy Cristina Álamo Suárez, Oier Ijurco, Yanisbel Ríos Laborde and Yoandra Chuen Gómez.

\bibliography{custom}

\clearpage

\appendix

\onecolumn
\section{Guidelines for Variation Adaptation}
\label{sec:appendix1}

\vspace{-0.20cm}

\begin{longtable}{p{0.2\textwidth}p{0.7\textwidth}}
 \\ \toprule
       \textbf{Language}  & \textbf{Adaptation guidelines} \\ \toprule

        Basque & Ataza honetan testu motz batzuk hizkuntza formal/estandarretik hizkuntza informalagora/euskalkietara berridatzi behar dira. Bakoitzak bere hizkuntza informal/dialektalean esango lukeen bezala idaztea da helburua. Hau horrela, ondorengo aldaketak proposatzen ditugu: 
            \begin{itemize}
            \vspace{-0.20cm}
                \item Esamolde edo hizkuntza informalagoa bilakatu.
                \vspace{-0.20cm}
                \item Ezaugarri dialektalak gehitu, bai lexiko aldetik eta bai gramatika edo fonetika aldetik. 
                \vspace{-0.20cm}
                \item Hika.
                \vspace{-0.20cm}
            \end{itemize}

        Erregistroa edo dialektoak barne hartzen dituen beste edozein aldaketa ongietorria da. Adibidez: 
        
        \vspace{-0.15cm}
        \paragraph{Jatorrizkoa:} Bi dantzari horiek dantza hunkigarria eskaini zuten herriko frontoian.
        \vspace{-0.15cm}
        \paragraph{Berridatzia:} Bi dantsari hoiek dantza emozionantia eskeiñi zuten herriko frontoien. \\ \midrule
        
        Spanish &  En esta tarea se deben reescribir algunos textos cortos del lenguaje formal/estándar a un lenguaje más informal/dialectal. El objetivo es adaptar las frases como cada persona lo diría en su propio lenguaje dialectal. De esta manera, se propone hacer los siguientes cambios:

            \begin{itemize}
            \vspace{-0.20cm}
                \item A nivel de registro: Más informal, reescribiéndola de manera más coloquial
                \vspace{-0.20cm}
                \item Con rasgos dialectales, sean léxicos, gramaticales o fonéticos
                \vspace{-0.20cm}
                \item Adaptar la ortografía para que refleje vuestra pronunciación, dialecto
                \vspace{-0.20cm}
            \end{itemize}

        Cualquier otro cambio que refleje un cambio de registro o dialecto es bienvenido.         Por ejemplo: 
        \vspace{-0.15cm}
        \paragraph{Frase original:} El amigo se quedó sin opciones cuando le dijeron que el autobús no pasaría más.
        \vspace{-0.15cm}
        \paragraph{Frase adaptada:} El socio se quedo botao cuando le dijeron que la guagua no pasaba ma.\\ \midrule

        English &  This task involves rewriting short texts from formal/standard language to a more informal/dialectal language. The objective is to rewrite the sentences as each person would say them in their own dialectal language. The following changes are proposed:

            \begin{itemize}
            \vspace{-0.20cm}
                \item At the register level: More informal, rewriting in a more colloquial manner
                \vspace{-0.20cm}
                \item With dialectal features, whether lexical, grammatical, or phonetic
                \vspace{-0.20cm}
                \item Adapting spelling to reflect your pronunciation, dialect
                \vspace{-0.20cm}
            \end{itemize}
        
        Any other changes that reflect a change in register or dialect are welcome.
        
        For example:
        \vspace{-0.15cm}
        \paragraph{Original phrase:} Everyone, hurry up now, dinner is about to get cold.
        \vspace{-0.15cm}
        \paragraph{Adapted phrase:} Y’all better hurry up now, supper’s fixin’ to get cold. \\ 
         
         \bottomrule
    \caption{Guidelines for standard to dialectal adaptations, both in Basque and Spanish, and an English translation}
    \label{tab:guidelines}
\end{longtable}

\section{Adaptation Process Information}
\label{sec:appendix2}

\subsection{Annotator Metadata}
\begin{table}[!ht]
\begin{small}
\centering
\makebox[\textwidth]{%
    \begin{subtable}{0.40\textwidth}
        \centering
        \begin{tabular}{llrr}
            \toprule
            \textbf{Variable} & \textbf{Category} & \textbf{N} & \textbf{\%} \\
            \toprule
            \textbf{Location}  & Gipuzkoa       & 7  &  58.34 \\
                               & Biscay      &  4 & 33.34  \\
                               & Navarre       &  1 & 8.34  \\ \midrule
            \textbf{Age}       & 20-30            & 5  &  41.67 \\
                               & 30-40            & 3  & 25.00  \\
                               & 40+              &  4  &  33.34 \\ \midrule
            \textbf{Gender}    & Male             &  5 &  41.67 \\
                               & Female           & 7  &  58.34 \\ \midrule
            \textbf{Background}  & Linguist         &  8 &  66.67 \\
                               & Non-linguist      & 4  & 33.34  \\
            \bottomrule
        \end{tabular}
        \caption{Demographic metadata of annotators. \textbf{N} = Count; \textbf{\%} = Percentage}
        \label{tab:demographics-eu}
    \end{subtable}

    \hspace{0.04\textwidth}%
    \begin{subtable}{0.40\textwidth}    
        \centering
        \begin{tabular}{llrr}
            \toprule
            \textbf{Variable} & \textbf{Category} & \textbf{N} & \textbf{\%} \\
            \toprule
            \textbf{Location}  &   Cuba     &  2 & 33.33  \\
                               &   Ecuador    &  1 & 16.67  \\
                               &   SpainAndalusia  & 1  &  16.67 \\
                               &   SpainTenerife  & 1  &  16.67 \\
                               &   Uruguay    & 1  &  16.67 \\ \midrule
            \textbf{Age}       & 20-30            &  1 &  16.67 \\
                               & 30-40            &  3 &  50.00 \\
                               & 40+              &  2  & 33.33  \\ \midrule
            \textbf{Gender}    & Male             & 2  & 33.34  \\
                               & Female           &  4 &  66.67 \\ \midrule
            \textbf{Background}  & Linguist         & 3  &  50.00 \\
                               & Non-linguist      &  3 &  50.00 \\
            \bottomrule
        \end{tabular}
        \caption{Demographic metadata of annotators. \textbf{N} = Count; \textbf{\%} = Percentage}
        \label{tab:demographics-es}
    \end{subtable}
}
\caption{Annotator metadata}
\end{small}
\end{table}

\subsection{Adaptation Type}

\begin{table}[!ht]
\centering
\begin{tabular}{lrrrr} 
\toprule
            & \multicolumn{2}{l}{\textbf{Basque}} & \multicolumn{2}{l}{\textbf{Spanish}} \\ \toprule
\textbf{Change type} & \textbf{N}           &\textbf{ \%}           & \textbf{N}            & \textbf{\%}          \\ \toprule
re-write    &       18      &     6.04         &   61           &   27.48           \\
dialectal   &     223        &     74.83         &      161        &      72.52        \\
Allocutive\_masc        &      37       &     12.41         &        -      &    -     \\  
allocutive\_fem    & 20    &  6.71   & -     & -   \\ \midrule

Total & 298 & & 222 & \\ \bottomrule

\end{tabular}
\caption{Number and percentage of change types in Basque and Spanish data. \textbf{N}: Count of examples; \textbf{\%}: Percentage}
\label{tab:change-type-statistics}
\end{table}

\subsection{Geographical Variants Distribution in Data}

\begin{figure}[H]
\begin{small}
    
    \centering
    \begin{subfigure}{0.40\textwidth}
        \centering
        \includegraphics[width=\textwidth]{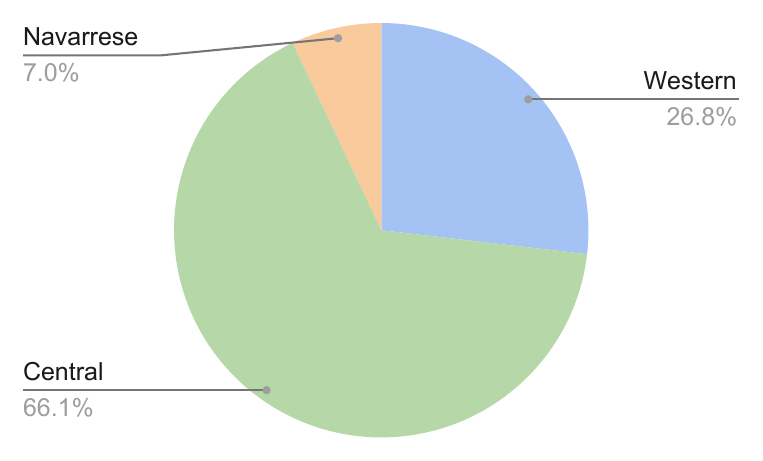}
        \caption{Number of examples per geographical variants in Basque}
        \label{fig:eu-dialect-labels}
    \end{subfigure}
    \hspace{1cm}
    \begin{subfigure}{0.40\textwidth}
        \centering
        \includegraphics[width=\textwidth]{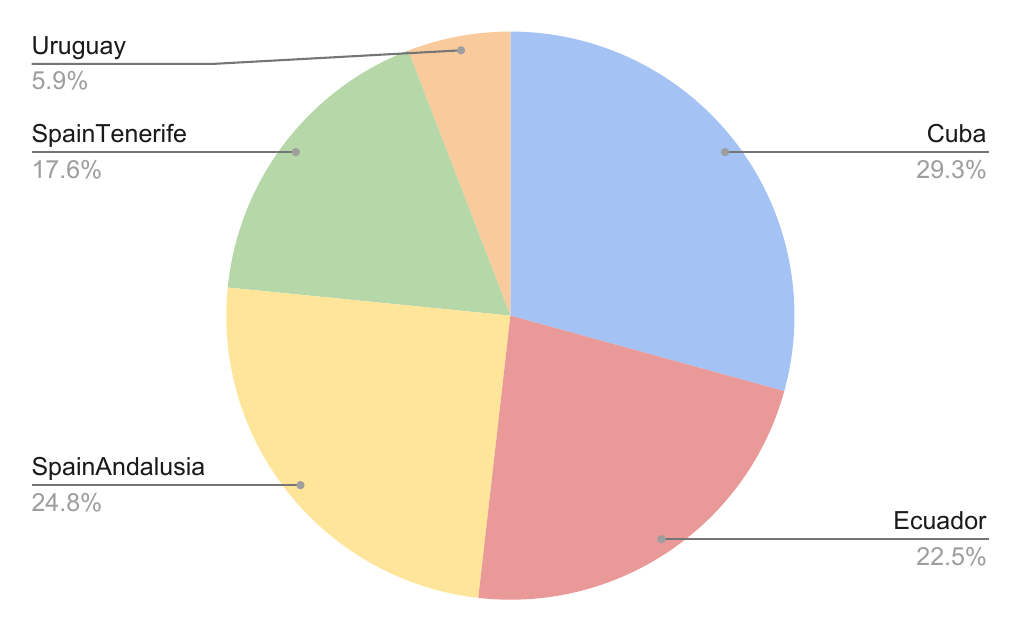}
        \caption{Number of examples per geographical variants in Spanish}
        \label{fig:es-dialect-labels}
    \end{subfigure}
\caption{Geographical variant label representation in XNLIvar}

\end{small}
\end{figure}


\clearpage
\onecolumn
\section{Prompts}
\label{sec:app_promp}

\begin{longtable}{p{0.2\textwidth}p{0.7\textwidth}}
 \\ \toprule
       \textbf{Task formulation}  & \textbf{Prompt} \\ \toprule

        nli-zero &  Please, answer in one word, with one of the following labels: <entailment>, <contradiction> or <neutral> Use exactly one of these three labels. \\ \midrule
        
        nli-few & "Say which is the inference relationship between these two sentences. Please, answer in one word, with one of the following labels: <entailment>, <contradiction> or <neutral> Use exactly one of these three labels. Here you have some examples: Postal Service were to reduce delivery frequency ->  The postal service could deliver less frequently: <entailment>. This elegant spa town on the edge of the Lac du Bourget has offered cures for rheumatism and other ailments for centuries -> The town was only established in the past fifty years: <contradiction>. And while we allow people to give a kidney to their child, we do not allow them to donate their heart -> You can't always donate organs to your child: <neutral>. \\ \midrule
        
        qa-zero & Are these two sentences entailed, contradicted or undetermined to each other? Please, answer in one word, with one of the following labels: <entailment>, <contradiction> or <neutral> Use exactly one of these three labels. \\ \midrule

        qa-few &  Are these two sentences entailed, contradicted or undetermined to each other? Please, answer in one word, with one of the following labels: <entailment>, <contradiction> or <neutral> Use exactly one of these three labels. Here you have some examples: Postal Service were to reduce delivery frequency -> The postal service could deliver less frequently: <entailment>. This elegant spa town on the edge of the Lac du Bourget has offered cures for rheumatism and other ailments for centuries -> The town was only established in the past fifty years: <contradiction>. And while we allow people to give a kidney to their child , we do not allow them to donate their heart -> You can't always donate organs to your child: <neutral>.\\ \midrule
        
        chain & You are an expert linguist and your task is to annotate sentences for the task of Natural Language Inference. This task consists in determining if a first sentence (premise) entails, contradicts or does not entail nor contradict the second sentence (hypothesis). Please, answer in one word, with one of the following labels: <entailment>, <contradiction> or <neutral> \textbackslash n Use exactly one of these three labels \textbackslash n Here you have a few examples:\textbackslash n Premise: Postal Service were to reduce delivery frequency. \textbackslash n Hypothesis: The postal service could deliver less frequently. \textbackslash n Answer: <entailment> \textbackslash n Premise: This elegant spa town on the edge of the Lac du Bourget has offered cures for rheumatism and other ailments for centuries. \textbackslash n Hypothesis: The town was only established in the past fifty years. \textbackslash n Answer: <contradiction> \textbackslash n Premise: And while we allow people to give a kidney to their child , we do not allow them to donate their heart. \textbackslash n Hypothesis: You can't always donate organs to your child. \textbackslash n Answer: <neutral> \\
         \bottomrule
    \caption{Different task formulation prompts for generative model prompting}
    \label{tab:prompts}

\end{longtable}

\clearpage
\onecolumn
\section{Basque Monolingual Discriminative Results}
\label{sec:monolingual-discriminative}

\begin{table}[!ht]    
\centering
    \begin{tabular}{lrrr} \toprule 
      & \multicolumn{3}{c}{\textbf{Translate-train}}  \\ \midrule
      & XNLIeu & XNLIeu\textsubscript{native} & XNLIeu\textsubscript{var} \\ \midrule
      RoBERTa-Euscrawl & 82.63  &  73.43  &  72.24  \\
      BERTeus &  78.15  &  68.81  &  63.67    \\ 
    \bottomrule 

    \end{tabular}
        \caption{Accuracy results for Basque monolingual discriminative experiments}
    \label{tab:discriminative-eu-monolingual}

\end{table}

\section{Additional Generative Results}
\label{sec:addicional-generative-results}

\begin{table}[!ht]
\begin{small}
    \centering
\begin{subtable}{\textwidth}
    \centering
    \resizebox{\textwidth}{!}{
    \begin{tabular}{lrrrrr|rrrrr} \toprule
     & \multicolumn{5}{c|}{\textbf{Llama-3.1-Instruct-8B}} & \multicolumn{5}{c}{\textbf{Gemma-2-it-9B}}  \\ \midrule
    
       &  nli-zero  &  nli-few  &  qa-zero  &  qa-few  &  chain &  nli-zero  &  nli-few  &  qa-zero  &  qa-few  &  chain  \\ \midrule
       XNLIeu & 20.30 & 	16.63 & 09.16 & 38.50	 & 51.76  & 55.61	 & 38.66	 &  37.96	 & 44.51 & 48.88 \\
       
      XNLIeu\textsubscript{native} & 21.36  & 17.90 & 07.05	
 & 36.24 & 41.83 & 61.19	 & 39.94	 & 41.55 & 	49.11 & 54.43 \\ 
      
      XNLIeu\textsubscript{var} & 20.45	 & 13.04 & 14.33  & 37.04	  & 46.22 & 53.47 & 39.04	 &  36.13	 & 41.72 & 45.53 \\ 
      \bottomrule
    \end{tabular}
    }
    \caption{Accuracy results with generative LLMs on Basque data.}
    \label{tab:generative-results-eu-additonal}
\end{subtable}       

\hfill

\begin{subtable}{\textwidth}
    \centering
    \resizebox{\textwidth}{!}{
    \begin{tabular}{lrrrrr|rrrrr} \toprule
     & \multicolumn{5}{c|}{\textbf{Llama-3.1-Instruct-8B}} & \multicolumn{5}{c}{\textbf{Gemma-2-it-9B}}  \\ \midrule
    
       &  nli-zero  &  nli-few  &  qa-zero  &  qa-few  &  chain &  nli-zero  &  nli-few  &  qa-zero  &  qa-few  &  chain  \\ \midrule
       XNLIes & 27.96	 & 	22.87 & 16.43 & 49.28 &  57.78  & 64.11  & 55.57	 & 44.73 & 57.41 &  66.83 \\
       
      XNLIeu2es\textsubscript{native} & 28.34  & 15.62 & 23.19
    & 48.79 & 62.80  & 69.24	 & 55.72	  & 50.24	 & 59.42  & 71.82  \\ 
      
      XNLIes\textsubscript{var} & 26.73 & 21.62 & 19.37  & 48.35  &  56.46 & 67.63 & 53.14  & 44.28 &  55.72	& 68.92 \\
      \bottomrule
    \end{tabular}
    }
    \caption{Accuracy results with generative LLMs on Spanish data.}
    \label{tab:generative-results-es-additional}
\end{subtable}
   
    \caption{Results with 8B and 9B LLMs.}
    \label{tab:generative-all-small}
\end{small}
\end{table}

\section{Per-dialect Accuracy Results}
\label{sec:accuracy-per-dialect}

\begin{table}[!h]
\begin{small}
\centering
 \resizebox{\textwidth}{!}{
    \begin{tabular}{lrrr|rrr|rrr} \toprule
     & \multicolumn{3}{c|}{\textbf{Model transfer}} & \multicolumn{3}{c|}{\textbf{Translate-train}} & \multicolumn{3}{c}{\textbf{Translate-test}} \\ \midrule
    
      &  Western  &  Central  &  Navarrese  &  Western  &  Central  &  Navarrese &  Western  &  Central  &  Navarrese \\ \midrule
      XLM-RoBERTa large &  71.25 & 67.17 & 71.43 & 73.75 &	74.45 & 71.43  & 71.67 & 72.42 & 74.60 \\ 
      mDeBERTa &  62.08 & 70.90 & 60.32 & 66.25 & 72.59 & 
      66.67 & 69.58 & 71.40 & 73.02 \\ \bottomrule

    \end{tabular}
    }

\caption{Accuracy results for discriminative models in Basque dialects}
\label{tab:accuracy-per-label-eu}

\end{small}
\end{table}

\clearpage
\begin{table}[!htbp]
\begin{small}

    \centering
    \resizebox{\textwidth}{!}{
    \begin{tabular}{lrrrrr} \toprule
     & \multicolumn{5}{c}{\textbf{Model transfer}}  \\ \midrule
    
       &  Cuba  &  Ecuador  &  SpainAndalusia  &  SpainTenerife  &  Uruguay  \\ \midrule
       XLM-RoBERTa large &   79.49 & 78.67 &	69.70 & 58.97 & 82.05   \\
      mDeBERTa &    76.92 & 78.67 & 70.30 & 58.97 & 79.49   \\ \toprule

    & \multicolumn{5}{c}{\textbf{Translate-train}}  \\ \midrule
    
       &  Cuba  &  Ecuador  &  SpainAndalusia  &  SpainTenerife  &  Uruguay  \\ \midrule
       XLM-RoBERTa large &  79.49	& 84.67 & 70.91 & 61.54 & 82.05   \\
      mDeBERTa &   73.33	& 81.33 & 69.09 & 53.85 & 87.18   \\ \toprule

        & \multicolumn{5}{c}{\textbf{Translate-test}}  \\ \midrule
    
       &  Cuba  &  Ecuador  &  SpainAndalusia  &  SpainTenerife  &  Uruguay  \\ \midrule
       XLM-RoBERTa large & 72.82 & 76.67 & 72.12 & 64.10 & 76.92   \\
      mDeBERTa &   69.23 & 77.33 & 67.88 & 62.39 & 79.49 \\ \bottomrule

    \end{tabular}
    }
    \caption{Accuracy results for discriminative modes in Spanish variants}
    \label{tab:accuracy-per-label-es-model-transfer}

\end{small}
\end{table}

\section{Spanish Correlation Between Adaptation Types and Accuracy}
\label{sec:spanish-analysis}

\begin{figure}[!ht]
    \centering
        \includegraphics[width=\textwidth]{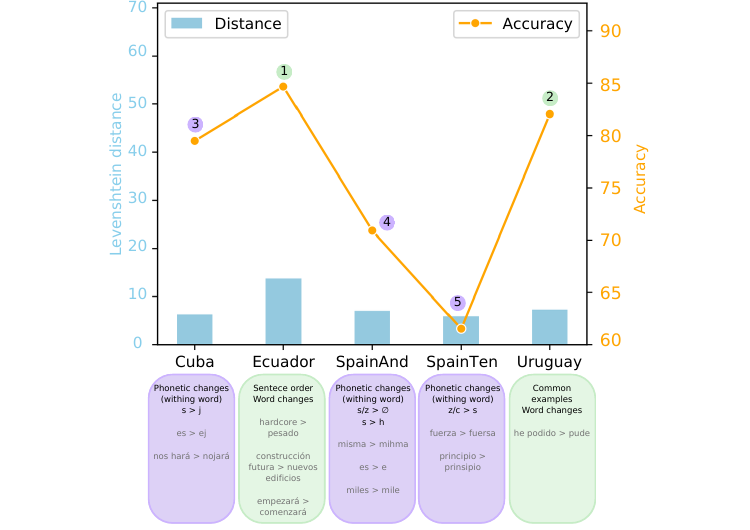}
        \caption{Spanish accuracy results and its correlation to types of linguistic adaptations. \colorbox{mygreen}{1} and \colorbox{mygreen}{2} have the highest accuracies, but changes usually involve word changes. For \colorbox{mypurple}{3}, \colorbox{mypurple}{4} and \colorbox{mypurple}{5}, the accuracy decreases respectively, as variations majorly involve phonetic changes.}
        \label{fig:esquema}
\end{figure}

\end{document}